\documentclass[10pt,twocolumn,letterpaper]{article}

\usepackage{wacv}
\usepackage{times}
\usepackage{epsfig}
\usepackage{graphicx}
\usepackage{amsmath}
\usepackage{amssymb}
\usepackage{multirow}
\usepackage[table,xcdraw]{xcolor}
\usepackage{makecell}

\usepackage[justification=centering, font={small}]{caption}

\wacvfinalcopy 

\def\wacvPaperID{****} 

\ifwacvfinal\fi
\begin{document}

\title{Safe Augmentation:\linebreak Learning Task-Specific Transformations from Data}

\author{Irynei Baran, Orest Kupyn\\
Ukrainian Catholic University\\
Lviv, Ukraine\\
{\tt\small \{i.baran, kupyn\}@ucu.edu.ua}
\and
Arseny Kravchenko\\
ods.ai\\
Minsk, Belarus\\
{\tt\small me@arseny.info}
}

\maketitle

\begin{abstract}
Data augmentation is widely used as a part of the training process applied to deep learning models, especially in the computer vision domain. Currently, common data augmentation techniques are designed manually. Therefore they require expert knowledge and time. Moreover, augmentations are dataset-specific, and the optimal augmentations set on a specific dataset has limited transferability to others. We present a simple and explainable method called \textbf{Safe Augmentation} that can learn task-specific data augmentation techniques that do not change the data distribution and improve the generalization of the model. We propose to use safe augmentation in two ways: for model fine-tuning and along with other augmentation techniques. Our method is model-agnostic, easy to implement, and achieves better accuracy on CIFAR-10, CIFAR-100, SVHN, Tiny ImageNet, and Cityscapes datasets comparing to baseline augmentation techniques. The code is available at \url{https://github.com/Irynei/SafeAugmentation}.

\end{abstract}


\section{Introduction}

Deep neural networks achieve human-level or even higher performance in many computer vision tasks, such as image classification, image restoration, image or video segmentation, etc. \cite{2018SurveyOnDeepLearning}. For example, the human top-5 image classification error on the ImageNet dataset is $5\%$, whereas the current state-of-the-art deep neural networks achieve nearly $3\%$ \cite{ImageNet_Challange}.

\begin{figure}[h!]
        \centering
        \includegraphics[width=0.9\linewidth]{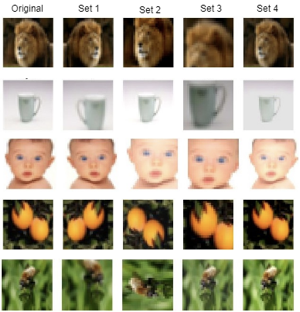}
        \caption[Example of using random subsets $s \subset S$ of safe augmentations on images from CIFAR-100 dataset]{
        Example of using random subsets $s \subset S$ of safe augmentations on images from CIFAR-100 dataset. Each transformation is applied with the probability $p=0.5$. \\ Each crops is of size $25x25$ pixels.
        }
        \captionsetup{justification=raggedright}
        \caption*{
        Set 1: HorizontalFlip, RandomContrast, RandomSizedCrop. \\ Set 2: RandomCrop, RandomContrast, RandomRotate90. \\ Set 3: RandomSizedCrop, RandomContrast, RandomCrop. \\Set 4: RandomContrast, RandomBrightness, RandomGamma.
        }
\label{fig:safe_augmnetation_examples}
\end{figure}

However, deep learning models require a massive amount of training data to be robust. Data augmentation is one of the approaches that can help to handle this issue by expanding training data using transformations that preserve semantic information and class labels. The choice of the data augmentation techniques to use for the specific dataset and task is not a trivial one. While some augmentations increase the performance and generalization of the model, others can have a negative impact. For instance, a horizontal flip is proven to be useful augmentation for ImageNet-like datasets, but not for the MNIST dataset 
\cite{lecun-mnisthandwrittendigit-2010}, as it changes the distribution of the data because horizontally flipped digits are often no longer valid digits.

Automated machine learning, in particular, automatic data augmentation is currently am important research topic. AutoAugment uses reinforcement learning to search for optimal augmentation policies along with the magnitudes and probabilities \cite{2018AutoAugment}. While these methods provide superior image classification results, they often have limited explainability and require lots of computational resources. Yet the majority of common data augmentation techniques are designed either empirically or by leveraging expert knowledge. Hence, it decreases the transferability of such methods between different tasks.

This paper aims to make another push on making the process of choosing data augmentations automatic and explainable, namely to learn from data which augmentation techniques lead to model generalization improvement. Our contributions are summarized below.
\begin{itemize}
    \item We introduce a simple, intuitive model-agnostic method for choosing augmentations that can be safely used during model training. In our implementation, we take a fixed set of the common image processing functions with default magnitudes and select a subset of augmentations which produce images that comes from the same distribution that existing ones while improving the accuracy of the main task(e.g., image classification, image segmentation).
    \item We propose two ways of using learned augmentations: for model fine-tuning and along with other augmentation techniques. Our experiments on different datasets and tasks show that Safe Augmentation works better than baseline augmentations and comparable with more advanced augmentation techniques while being intuitive, explainable, and straightforward.
\end{itemize}


\section{Related Work}


\subsection{Data Augmentation}

The recent paper by Hernández-García and König has shown that data augmentation alone can achieve the same or even higher performance than explicit regularization techniques (weight decay \cite{1991WeightDecay}, dropout \cite{2014Dropout}, etc.), without wasting model capacity \cite{2018AugmentationInsteadOfRegularization}.

\textbf{Traditional augmentation}. The most common type of data augmentations are geometric transformations, such as flipping, cropping, rotating, scaling, etc., and color transformations, such as adjusting color, brightness, resolution, etc. 
They are often called generic or traditional augmentations. They all fall under the category of data warping and are usually performed in the data space, e.g., Wong \etal have shown that it is more efficient to perform
data augmentation in data space than in feature space as long as label preserving transforms are known \cite{2016WhenToWarp}. This type of transformations is easy to use and efficient to implement. One main disadvantage is that you need to have expert knowledge in the image domain to choose transformations that will not affect the correctness of the image labels.
Traditional augmentations are broadly used and have shown excellent results in reducing overfitting and improving model performance \cite{EffectivenessAugm2017} \cite{2017GenericDataAugmentation}.

\textbf{Generative Adversarial Networks (GANs)}. In 2014 Goodfellow \etal proposed a new class of neural networks that can generate realistic data from scratch using generator and discriminator networks that are trained in the minimax two-player game framework \cite{2014GAN}. GANs can be used as a form of unsupervised data augmentation by generating new data from the source distribution.
They have also been used for style transfer, e.g., transferring images from one weather condition to another. These generated images can be used to help the model to work in different conditions, for instance, to train autonomous cars to drive in night or snow, having collected data from sunny weather only. GANs are also shown to be successfully used for data augmentation in the medical imaging domain by synthetically augment mammogram and MRI images \cite{2018BrainSegmentation, 2018Mammogram}.

\subsection{Automated Data Augmentation}
Despite described advantages, common data augmentation methods are usually dataset-specific and designed manually, which require prior expert knowledge and time. Recently, many researches were focused on the automation of the process of data augmentation. We divide them into the following two groups.

\textbf{Generate augmented data directly}.
Smart Augmentation proposed by Lemley, Bazrafkan, and Corcoran can automatically generate augmented data by merging two or more samples from the same class, in a way that reduces the loss of the original model \cite{2017SmartAugmentation}. DeVries and Taylor proposed a domain-independent data augmentation technique by using simple transformations in the learned feature space. They train a sequence autoencoder to construct a learned feature space in which they extrapolate between samples \cite{2017AugmentationFeatureSpace}. Tran \etal. introduced a novel Bayesian method for generating additional data based on the distribution learned from the training set \cite{2017BayesianDataAugmentation}. Generative adversarial networks have been extensively used for producing augmented data. For example, Antoniou, Storkey, and Edwards presented DAGAN - An image conditional GAN-based model that learns from one data item how to generate other realistic within-class data items. DAGAN can be applied to unseen classes of data and can also enhance few-shot learning systems.
\cite{2017AugmentationGAN}. Another approach called DADA: Deep Adversarial Data Augmentation was proposed by Zhang \etal to train deep learning models in extremely low data regimes. They show that that DADA outperforms both traditional data augmentation and a few GAN-based options \cite{2018DADA}.

\begin{figure*}[h!]
        \centering
        \includegraphics[width=\linewidth]{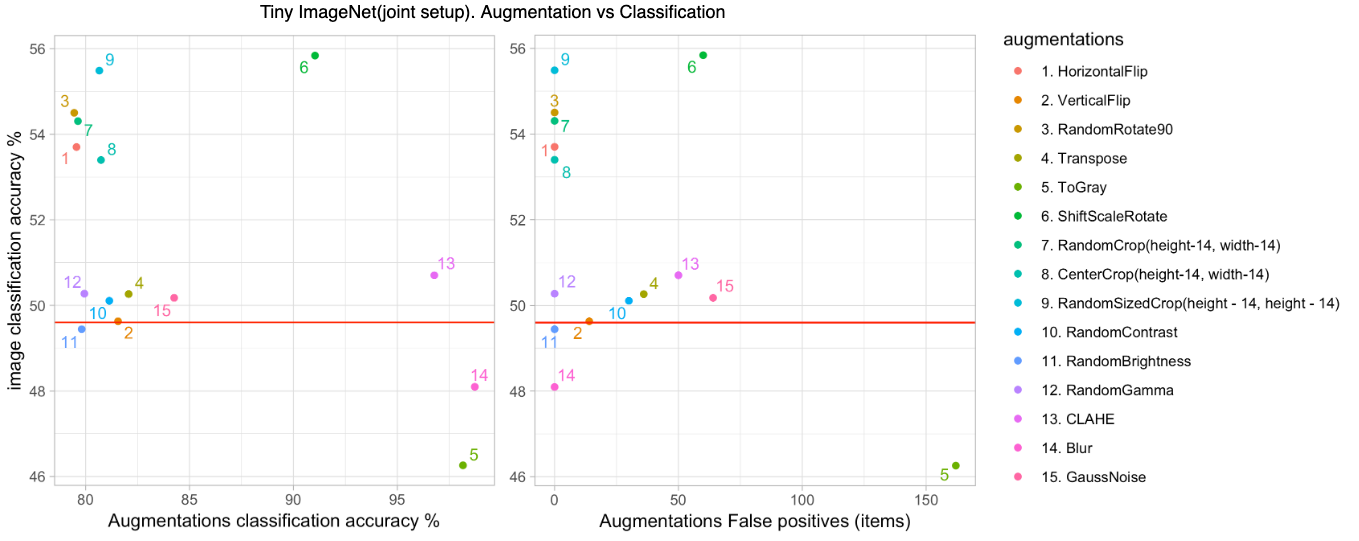}
\caption{Image classification vs Augmentation false positives (step 2) and Augmentation classification (step 3). \\ Red line denotes image classification accuracy without augmentations.}
\label{fig:safe_augmnetation_metrics}
\end{figure*}

\textbf{Generate data transformations}. 
Ratner \etal learned generative sequence model over user-defined transformations using GAN-like framework. Their idea is to compose and parameterize a set of user-specified transformation functions in ways that are diverse but still preserve class labels. Their approach allows leveraging domain knowledge flexibly and straightforwardly. \cite{2017ComposeDomain-SpecificTransformations}.

Cubuk \etal proposed a new procedure called AutoAugment\cite{2018AutoAugment} to learn augmentation policies that lead to the highest accuracy of the image classification model on a given dataset. They created a search space of data augmentation policies and used a search algorithm based on reinforcement learning to find the optimal one. The results are great: they achieved state-of-the-art accuracy on CIFAR-10, CIFAR-100, SVHN, and ImageNet datasets. Moreover, It is shown that policies learned from one dataset can be transferred to other similar datasets. One of the drawbacks of AutoAugment is the computational complexity and long training time due to the extensive search space of possible policies \cite{2018AutoAugment}. Many interesting works have been published trying to address this issue, including Fast AutoAugment \cite{2019FastAutoAugment} and PBA (Population Based Augmentation) \cite{2019PBD}, which achieved similar results to AutoAugment but using much more efficient algorithms.


\section{Proposed Method}

We present an intuitive approach for learning data transformations that can be safely used during the model training. Our learned set of augmentations is called safe augmentations, which can be used either for model fine-tuning or along with other augmentation techniques.

Our method does not require substantial computational resources and can be easily performed along with the main task.

\subsection{Learning Safe Augmentations}
We propose to learn safe augmentations from data using an arbitrary convolutional neural network (CNN). Consider a dataset $D$ and a set of all available augmentation techniques $A$. The task is to define which transformations from set $A$ do not change the distribution of the $D$, i.e., to select $S \subset A$, where $S$ is a set of safe augmentation. Our pipeline can be divided into four main steps.

\begin{itemize}

\item \textbf{Step 1.} Train the CNN to solve the following multi-label classification problem. Given a set $A$, for every batch of images, a random subset $a \subset A$ is applied. The subset $a$ is of random size from $0$ to the defined maximal size. In our experiments, we used maximum size of 5. Each transformation from subset $a$ is applied with the probability $p=1$. The model tries to predict which augmentations were applied. As a loss function, we use $L_{augm}$ - a multi-label one-versus-all loss based on max-entropy, between input $x$ and target $y$. $L_{augm}$ is equivalent to applying $sigmoid$ function along with the binary cross-entropy loss.

\item \textbf{Step 2.} After the model is being trained, evaluate it on the unseen test data without any augmentations and collect per-label false positives, i.e., how many times the model predicts that the specific augmentation technique was applied when, in fact, it was not.

\item \textbf{Step 3.} Evaluate the model on the unseen test data using the same procedure of applying augmentations as in the training phase. Collect per-label classification accuracy for each transformation technique.

\item \textbf{Step 4.} Divide all augmentations into two groups: safe and others. If the model fails to distinguish whether a particular transformation was applied and the transformation is never predicted on the clean set, then this transformation does not change the data distribution and can be safely used during the training of the original task. Thus, we consider augmentation as safe if it has relatively low per-label classification accuracy on the test set with augmentations (step 3) and low false positive rate on the clean test set without augmentations (step 2).
 
\end{itemize}

\textbf{Example}. Figure \ref{fig:safe_augmnetation_metrics} shows the described above metrics for the Tiny ImageNet dataset along with the image classification accuracy of every single augmentation. $Blur$ is an example of non-safe transformation with the low false positive rate on the clean test set and very high augmentation classification accuracy on the augmented test set, so the model can predict when $Blur$ was applied. On the other hand, $HorizontalFlip$ is an example of safe augmentation with both low false positive rate and augmentation classification accuracy. It is clearly shown than $HorizontalFlip$ significantly increases the image classification accuracy on the Tiny ImageNet dataset, whereas $Blur$ decreases it.

Note that we can only evaluate one transformation at a time, i.e., we cannot take into account the impact of different augmentations on each other. So the combination of safe augmentations is not necessarily safe. For example, given a dataset of $32 x 32$ images and image classification task, our method found that \emph{RandomCrop(25, 25)} and \emph{CenterCrop(25, 25)} are safe augmentations. However, when these two functions are applied together, it is likely that such a combination is no longer safe because the augmented image could be too small.


\subsection{Joint Learning}

To learn augmentations that not only do not change the data distribution but also improve original task accuracy(e.g., image classification, image segmentation), we trained the multi-label classification problem (step 1) in a joint learning setup. To do that, we propose to modify the architecture of the original models in the following way.

For the image classification task, the new loss $L_{total}$ is calculated as sum of the augmentation classification loss $L_{augm}$ and the image classification loss $L_{class}$.
\begin{equation}
    L_{total} = L_{augm} + L_{class}
\end{equation}
where $L_{class}$ is the cross-entropy loss.

For the semantic image segmentation task, the new loss $L_{total}$ is calculated in a similar way as a sum of the augmentation classification loss $L_{augm}$ and the semantic segmentation loss $L_{segm}$.
\begin{equation}
        L_{total} = L_{augm} + L_{segm}
\end{equation}

where $L_{segm}$ is the cross-entropy loss same as $L_{class}$, but here $x$ is a two-dimensional predicted mask, $y$ is the two-dimensional  target mask, and the goal is to label every pixel in $x$ with the correct class.

For each augmented batch of images, we calculate defined above $L_{total}$ loss and then perform gradient updates. Joint learning setup helps find a better set of safe augmentations that can be used to improve the performance of the original task. All the results presented were obtained using this approach.

\subsection{Using Safe Augmentations}
Having learned the set of safe augmentations $S$ for a given dataset and task, we propose to use them in the following two ways:
\begin{itemize}
\item \textbf{For model fine-tuning}

 \textbf{Step 1.} Train the original task using a set of all augmentations $A$. For every batch of images, a random subset $a \subset A$ of fixed size is applied. Each transformation is applied with the fixed probability $p=0.5$.

\textbf{Step 2.} Fine-tune the already pre-trained model on all augmentations using a subset of safe augmentations $S$. For every batch of images, a random subset $s \subset S$ of fixed size is applied. The subset size and probability of applying transformations are the same as in the previous step.

We believe that using all augmentations, including those that change the data distribution can force the model to learn more general features. Thus, they can be used for the model pre-training. After that, we need to fine-tune the model using safe augmentations for learning dataset-specific features.

\item \textbf{Along with other augmentation techniques}

Safe augmentations alone cannot provide enough generalization.
In our experiments, we show that using safe augmentations along with other augmentation techniques, e.g., baseline augmentations and Cutout \cite{2017cutout} leads to better results.
\end{itemize}

\begin{figure*}[h!]
        \centering
        \includegraphics[width=\linewidth]{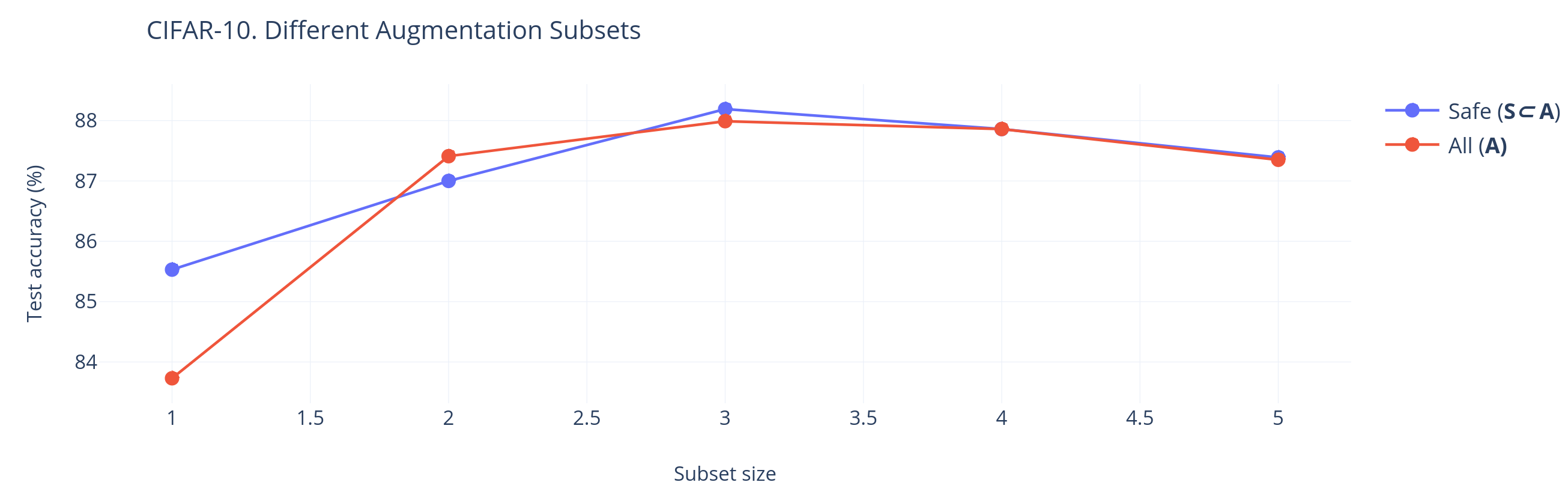}
        \caption{
        Evaluation of different augmentation subsets on CIFAR-10}
\label{fig:ablation-study}
\end{figure*}

\section{Experimental Evaluation}

\subsection{Implementation Details}
We implemented all of our models using PyTorch \cite{pytorch} \cite{paszke2017automatic}.
All  models were trained on a single GTX 1080 GPU. We perform the phase of learning safe augmentations using the joint learning setup.

\textbf{Augmentations.}
We use a main set $A$ of $15$ common augmentations (see Table \ref{tab:safe-augmentations}) from \textit{albumentations} library that provides fast image transforms based on highly-optimized OpenCV library \cite{2018Albumentations}.

\textbf{Image classification.}
 As the main model, we use DenseNet-121 \cite{densenet}. Both learning safe augmentations and image classification tasks are trained from scratch using stochastic gradient descent (SGD) optimizer using the batch of size 256. The initial learning rate is set to $10^{-1}$, momentum to $0.9$ and weight decay to $0.0005$. All models are trained for $500$ epochs with reducing learning rate on the plateau by $0.1$ with $10$ epochs patience and early stopping with $20$ epochs patience.

\textbf{Image segmentation.}
As the main model, we use Feature Pyramid Network (FPN) \cite{FPN} with the DenseNet-121 \cite{densenet} backbone. Both learning safe augmentations and image classification tasks are trained from scratch using Adam \cite{ADAM} optimizer.  The initial learning rate is set to $10^{-4}$. All models are trained for $200$ epochs with reducing learning rate on plateau by $0.5$ with $7$ epochs patience and early stopping with $15$ epochs patience.

\subsection{Augmentation Subset Size}
We investigate impact of  different subset sizes on image classification accuracy (see Figure  \ref{fig:ablation-study}). Empirically proven that subset of size 3 leads to the best image classification accuracy for both all augmentations set $A$ and safe augmentations set $S$. Hence, in all our experiments we are using $a \subset A$ and $s \subset S$ of size 3.

\begin{table}[h]
\begin{center}
    \begin{tabular}{l | c | c | c | c}
    & C-10 & C-100 &  Tiny INet & SVHN \\
       \hline
    HorizontalFlip & \checkmark & \checkmark & \checkmark & \\
        \hline
    VerticalFlip  & & &  & \\
        \hline
    RandomRotate90  & \checkmark & \checkmark & \checkmark & \\
        \hline
    Transpose   & & &  &  \\
        \hline
    ToGray  & & &  &  \\
        \hline
    ShiftScaleRotate  & & &  &  \\
        \hline
    RandomCrop & \checkmark & \checkmark & \checkmark & \\
        \hline
    CenterCrop  & \checkmark & \checkmark & \checkmark & \\
        \hline
    RandomSizedCrop  & \checkmark & \checkmark & \checkmark & \\
        \hline
    RandomContrast  & & \checkmark &  & \checkmark \\
        \hline
    RandomBrightness  & \checkmark & \checkmark & \checkmark & \checkmark \\
        \hline
    RandomGamma  & \checkmark & \checkmark & \checkmark & \checkmark \\ 
        \hline
    CLAHE   & & &  & \\
        \hline
    Blur  & & &  &  \\
        \hline
    GaussNoise  &  &  &  & \checkmark \\
    \hline
    \end{tabular}
\end{center}
\caption {Safe augmentations found using joint learning setup for CIFAR-10, CIFAR-100, Tiny ImageNet and SVHN datasets.} \label{tab:safe-augmentations}
\end{table}

\subsection{Quantitative Evaluation on Image Classification}

We evaluate our method on 4 popular image classification datasets, namely CIFAR-10 \cite{Cifar-10}, CIFAR-100 \cite{Cifar-100}, SVHN \cite{SVHN} and Tiny ImageNet. For every dataset, we train the image classification task using different augmentation techniques and models. All augmentations in our image classification experiments are applied with the fixed probability $p=0.5$ and with default magnitude.

\begin{table*}[h!]
\centering
\begin{tabular}{l c c c c c c}
\multirow{2}{*}{}                                            & \multicolumn{2}{c|}{CIFAR-10}   & \multicolumn{2}{c|}{SVHN}       & \multicolumn{1}{c|}{CIFAR-100} & Tiny ImageNet \\ \cline{2-7} 
 & \multicolumn{1}{c|}{\makecell{DenseNet \\ 121}} & \multicolumn{1}{c|}{\makecell{DenseNet \\ 169}}    & \makecell{DenseNet \\ 121} & \multicolumn{1}{|c|}{\makecell{DenseNet \\ 169}}    & \multicolumn{1}{c|}{\makecell{DenseNet \\ 121}} & \makecell{DenseNet \\ 121}  \\ \hline
Without                 & 79.39          & 79.52          & 95.87          & 96.19          & 53.55     & 49.60        \\ \hline
Baseline                 & 87.31          & 87.15          & 95.99          & 95.63         & 61.67      & 49.73          \\ \hline
Safe                  & 87.85          & 86.84          & 96.14          & 96.67          & 64.17    & 57.61       \\ \hline
All                     & 87.79          & 88.15          & 96.19          & 96.43          & 65.82    & 58.66       \\ \hline
Fine-tuned on All     & 88.68          & 88.15          & 96.19          & 96.58          & 65.83  & 58.65          \\ \hline
Fine-tuned on Safe     & 88.38          & 88.21          & \textbf{96.36} & \textbf{97.01} & 65.93   & 58.85           \\ \hline
Fine-tuned on Safe v2* & \textbf{88.59} & \textbf{88.46} & -              & -              & \textbf{65.99} &  \textbf{59.00}\\ \hline
\end{tabular}
\caption {Test top-1 accuracy (\%). All results are averaged over 3 runs. \\  All fine-tuned experiments were performed using models pre-trained on all augmentations. \\ * Safe v2 is defined manually by removing RandomCrop and CenterCrop.}
\label{tab:safe-results}
\end{table*}

\begin{itemize}
\item \textbf{CIFAR-10 and CIFAR-100.}
    We used all crops described in the Table \ref{tab:safe-augmentations} with the size $(25 x 25)$.
    As a baseline augmentations, we used horizontal flips with 50 \% probability, zero-padding and random crops, which are conventional for these datasets \cite{cifar_baseline}.
    The training data is also normalized by the respective dataset statistics.
    
    Our method found almost the same set of safe augmentations for CIFAR-10 and CIFAR-100 datasets,  which makes sense because these datasets are very similar (see Table \ref{tab:safe-augmentations}). We manually defined another set called Safe v2 by removing \textit{RandomCrop} and \textit{CenterCrop}, since our approach can only evaluate transformations independently.
    
    Fine-tuning using set Safe v2 achieves the best accuracy both on CIFAR-10 and CIFAR-100 comparing to other evaluated augmentation approaches, namely without augmentations, baseline, only safe, only all and different fine-tuned setups (see Table \ref{tab:safe-results}).

\item \textbf{SVHN.}
    Same as for CIFAR-10 and CIFAR-100 we used all crops described in the Table \ref{tab:safe-augmentations} with the size $(25 x 25)$.
    As a baseline augmentations, we used zero-padding and random crops. The training data is normalized by the respective dataset statistics.
    
    Our method found four safe augmentations (see Table \ref{tab:safe-augmentations}), which all are color-based transformations and can be safely used with digits.
    
    Fine-tuning using Safe set produces the best accuracy both for DenseNet-121 and DenseNet-169 models comparing to other evaluated augmentation approaches (see Table \ref{tab:safe-results}).

\item \textbf{Tiny ImageNet.}
    We used all crops described in the Table \ref{tab:safe-augmentations} with the size $50 x 50$.
    As a baseline augmentations, we used horizontal flips with
    $p=0.5$  and random distortions of colors, which are a standard data augmentation techniques for the ImageNet dataset \cite{imagenet_cvpr09, imagenet_baseline}. 
    
    Our method found the same safe augmentations (see Table \ref{tab:safe-augmentations}) as for CIFAR-10. We manually defined another set called Safe v2 in the same way as for  CIFAR-10 and CIFAR-100.
    
    Fine-tuning using Safe set again achieves the best accuracy comparing to other evaluated augmentation approaches (see Table \ref{tab:safe-results}).
    
\end{itemize}

\textbf{Safe Augmentation vs AutoAugment.}
We also evaluated our second proposed way of using safe augmentations along with other augmentation techniques. We compare our approach with AutoAugment \cite{2018AutoAugment} and Cutout \cite{2017cutout} on CIFAR-10, CIFAR-100 and SVHN datasets. For CIFAR-10 and CIFAR-100 we use cutout of size $16x16$ and for SVHN of size $20x20$ the same way as AutoAugment does. Safe augmentations are applied along with the baseline and Cutout.

The results on Table \ref{tab:safe-results-similar-methods} show that using safe augmentations in both fine-tuning way and along with other augmentation techniques lead to better results.  

\subsection{Quantitative Evaluation on Image Segmentation}
\begin{table}[h]
\begin{center}
    \begin{tabular}{l}
     Augmentations  \\
    \hline
    HorizontalFlip \\
    RandomBrightness \\
    RandomGamma \\
    Transpose \\
    \end{tabular}
\end{center}
\caption {Safe augmentations found using joint learning setup for the Cityscapes \cite{Cityscapes} dataset.} \label{tab:safe_cityscaps}
\end{table}

We also evaluate our method on Cityscapes - a popular image segmentation dataset \cite{Cityscapes}. We train both augmentation classification task and image classification task using the batch of size $16$. We rescale every image to $256x256$ pixels due to the limited training resources.
\begin{table}[h!]
\centering
\begin{tabular}{l c }
                   & FPN(DenseNet-121) \\ \hline
Without            &  45.34       \\ \hline
Baseline          & 51.11         \\ \hline
Safe             &  51.58    \\ \hline
All & 59.41 \\ \hline
Fine-tuned on All  & 60.37 \\ \hline
Fine-tuned on Safe & \textbf{62.09}\\ \hline
\end{tabular}
\caption {Validation IoU(\%) on the Cityscapes \cite{Cityscapes} dataset \\ on single FPN model with DenseNet-121 backbone}.
\label{tab:safe-results-cityscapes}
\end{table}
\begin{table*}[h!]
\centering
\begin{tabular}{l c c c c c}
                   & \multicolumn{2}{c|}{CIFAR-10}                       & \multicolumn{2}{c|}{SVHN}                                            & CIFAR-100      \\ \cline{2-6} 
\multirow{-2}{*}{} & \multicolumn{1}{c|}{\makecell{DenseNet \\ 121}}   & \multicolumn{1}{c}{\makecell{DenseNet \\ 169}} & \multicolumn{1}{|c|}{\makecell{DenseNet \\ 121} }                  & \multicolumn{1}{c|}{\makecell{DenseNet \\ 169} }   & \makecell{DenseNet \\ 121}   \\ \hline
Without            & 79.39          & 79.52                      & \multicolumn{1}{c}{95.87}                         & 96.19          & 53.55          \\ \hline
Best Policy AA     & 84.60          & 85.51                      & \multicolumn{1}{c}{96.67}                         & 96.65          & 59.34          \\ \hline
Baseline           & 87.31          & 87.15                      & \multicolumn{1}{c}{95.99}                         & 95.63          & 61.67          \\ \hline
Baseline + Cutout\             & 88.10          & 88.84                     & \multicolumn{1}{c}{96.31}                         & 96.39          & 63.26          \\ \hline
\rowcolor[HTML]{E5F7E5} 
{\color[HTML]{000000} Finetuned Safe*}                             & {\color[HTML]{000000} 88.59} & {\color[HTML]{000000} 88.46}  & {\color[HTML]{000000} 96.36} & {\color[HTML]{000000} \textbf{97.01}} & {\color[HTML]{000000} 65.99} \\ \hline
\rowcolor[HTML]{E5F7E5} 
\begin{tabular}[c]{@{}l@{}}Safe* + Baseline + Cutout\end{tabular} & 88.16                        & 88.32                                              & 96.73                        & 96.39                                 & 65.39                        \\ \hline
AutoAugment                 & \textbf{90.77} & \textbf{90.58}   & \multicolumn{1}{c}{\textbf{96.76}}                & 96.66 & \textbf{68.70} \\ \hline
\end{tabular}
\caption {Test top-1 accuracy (\%). Comparison with Cutout \cite{2017cutout} and AutoAugment \cite{2018AutoAugment}. \\ 
Safe* means \textit{Safe} for SVHN and \textit{Safe v2} for CIFAR-10 and CIFAR-100}.
\label{tab:safe-results-similar-methods}
\end{table*}
We also change the size of the crops to $512x512$. All augmentations in our experiments with the image classification task are applied with the fixed probability $p=0.5$ and with default magnitude.

\begin{itemize}
    \item \textbf{CityScapes}. As a baseline augmentations, we used horizontal flips with $p=0.5$ and rotation with $p=0.5$ with angle chosen randomly from 0 to 20 degrees, Our method found four safe augmentations (see Table \ref{tab:safe-augmentations}).
    
    Fine-tuning using Safe set again achieves the best accuracy comparing to other evaluated augmentation approaches (see Table \ref{tab:safe-results-cityscapes}).
\end{itemize}

\section{Conclusion}
This paper introduces Safe Augmentation, a simple yet efficient algorithm for automatic selection of data augmentations with promising results on quantitative benchmarks. In addition to the simplicity, explainability and flexibility \b{Safe Augmentation} also introduce two different ways of using learned augmentations along with other augmentations techniques. We plan to extend Safe Augmentation further for automatic parameter selection, as well as for adversarial robustness.
\section{Acknowledgement}
The authors were supported by the WANNABY, SoftServe and Ukrainian Catholic University. We thank ods.ai community for the constructive feedback, advice and ideas.
{\small
\bibliographystyle{ieee}
\bibliography{egpaper_final}
}

\end{document}